\documentclass{article}

\usepackage{arxiv}

\usepackage[utf8]{inputenc} 
\usepackage[T1]{fontenc}    
\usepackage{hyperref}       
\usepackage{url}            
\usepackage{booktabs}       
\usepackage{amsfonts}       
\usepackage{nicefrac}       
\usepackage{microtype}      
\usepackage{lipsum}
\usepackage{fancyhdr}       
\usepackage{graphicx}       
\graphicspath{{media/}}     

\usepackage{amssymb}
\usepackage{lscape} 
\usepackage{amsmath}
\usepackage{float}
\usepackage{subcaption}
\usepackage{multirow}
\usepackage{tabularx}
\usepackage{comment}
\usepackage{xspace}
\usepackage{makecell}
\usepackage{enumitem}
\usepackage{tikz}
\usepackage{pifont}
\usepackage{algorithmic}
\usepackage{textcomp}
\usepackage{booktabs}
\usepackage{color}
\usepackage[noabbrev,capitalize]{cleveref}

\usepackage[numbers]{natbib} 
\hypersetup{
    colorlinks=false, 
    pdfborder={0 0 0}, 
}

\newcommand{\ie}{\mbox{i.e.}\xspace}
\newcommand{\eg}{\mbox{e.g.}\xspace}
\newcommand{\mae}{\mbox{MAE}\xspace}

\title{Generic Multi-modal Representation Learning for Network Traffic Analysis}

\author{
  Luca Gioacchini, Marco Mellia \\
  Politecnico di Torino \\
  Italy\\
  \texttt{first.last@polito.it} \\
   \And
  Idilio Drago \\
  Università di Torino \\
  Italy\\
  \texttt{idilio.drago@unito.it} \\
   \And
  Zied Ben Houidi, Dario Rossi \\
  DataCom Lab, Huawei Technologies Co. Ltd \\
  France\\
  \texttt{first.mid.last@huawei.com} \\
}

\begin{document}
\maketitle

\begin{abstract}
Network traffic analysis is fundamental for network management, troubleshooting, and security. Tasks such as traffic classification, anomaly detection, and novelty discovery are fundamental for extracting operational information from network data and measurements. We witness the shift from deep packet inspection and basic machine learning to Deep Learning (DL) approaches where researchers define and test a custom DL architecture designed for each specific problem.
We here advocate the need for a general DL architecture flexible enough to solve different traffic analysis tasks. We test this idea by proposing a DL architecture based on generic data adaptation modules, followed by an integration module that summarises the extracted information into a compact and rich intermediate representation (i.e., embeddings). The result is a flexible Multi-modal Autoencoder (MAE) pipeline that can solve different use cases. We demonstrate the architecture with traffic classification (TC) tasks since they allow us to quantitatively compare results with state-of-the-art solutions. However, we argue that the MAE architecture is generic and can be used to learn representations useful in multiple scenarios. 
On TC, the MAE performs on par or better than alternatives while avoiding cumbersome feature engineering, thus streamlining the adoption of DL solutions for traffic analysis.
\end{abstract}


\section{Introduction}
\label{s:intro}

The abundance of methods to collect data and the availability of simple frameworks sparked the adoption of Machine Learning (ML) and Deep Learning (DL) algorithms to solve network traffic analysis problems. Examples are seen in traffic classification~\citep{rezaei2019deep,pacheco2019towards,horowicz2022few,aceto2019mobile}, in network security problems~\citep{sabir2021machine,ferrag2020deep,mishra2019detailed}, and in the unsupervised exploration of traffic~\citep{ring2017ip2vec,gioacchini2021darkvec,kallitsis2021zooming}, to cite a few. 

In this context, researchers and practitioners have started applying solutions developed by the AI community to solve various tasks~\citep{rezaei2019deep,ferrag2020deep}, sometimes borrowing Computer Vision (CV) methods~\citep{horowicz2022few,wang2017endend}, sometimes Natural Language Processing (NLP) methods~\citep{ring2017ip2vec,gioacchini2023idarkvec,cohen2020dante}.

However, the application of such algorithms in traffic analysis problems is often done by artificially forcing the input data and measurements to fit the selected method. Examples include creating an ``artificial image'' out of univariate time series to apply a 2D-Convolutional Neural Network (CNN)~\citep{horowicz2022few}; or mixing unrelated features in a single image-like matrix~\citep{wang2017endend}. In addition, the DL models are often complex and need to be tuned (in an \textit{end-to-end} fashion) for each input data type and target problem, to learn salient features and solve the specific problem at hand.  

This raises the question of whether we can find a simple yet efficient Deep Neural Network (DNN) that is generic enough to seamlessly encode different traffic measurements and solve multiple traffic analysis problems. The key idea is to let the general DL architecture produce a compact representation (or embeddings) of the often diverse and humongous input data. These embeddings could then be employed to solve other specific final problems (or tasks) without the burdens of building models from scratch starting from the raw features and measurements.

We here propose the adoption of a generic DL architecture to seamlessly create intermediate embeddings that succinctly capture the salient features of raw traffic data. 
A key question is how to properly learn such embeddings from the different types of input data typically found in traffic measurements. Network data consists of two modalities (see our preliminary work at~\citep{houidi2022towards}): (i) \textit{entities} (or categorical features) like IP addresses and port numbers; and (ii) \textit{quantities}, or various real-valued measurements like packet and flow size, time, etc. 
Here we extend this vision.
First, we adopt state-of-the-art approaches~\cite{gioacchini2023idarkvec} to learn representations from \textit{sequences of entities}, while we let general DL architectures handle packet payload and generic traffic statistics.
We then use a Multi-modal Autoencoder (MAE) to integrate and summarise each representation into a common and compact embedding space. Notice that the MAE is trained using a self-supervised task and thus enables the generation of the embeddings independently of the final downstream task. 

\begin{figure}[!t]
    \centering
    \includegraphics[width=\linewidth]{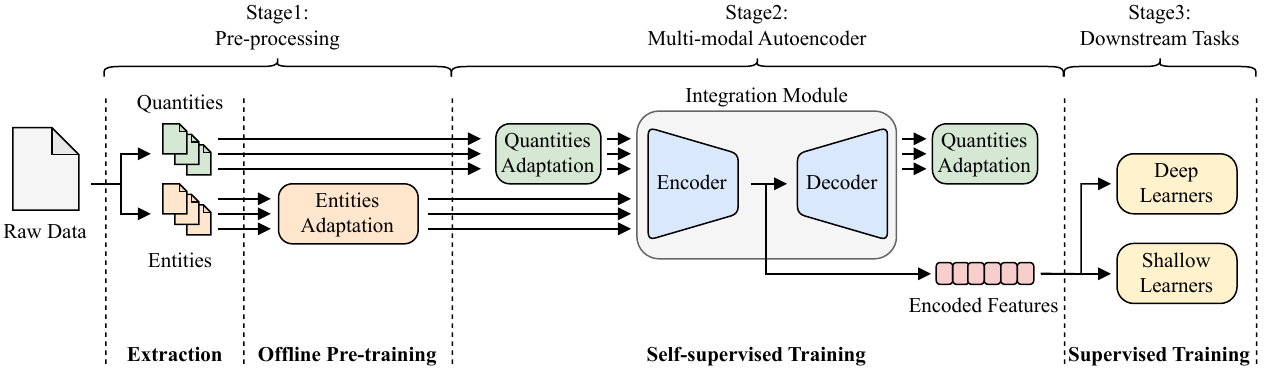}
    \caption{Overview of the modular Machine Learning pipeline relying on \mae.}
    \label{fig:pipeline}
\end{figure}

In \cref{fig:pipeline} we provide the overview of the full pipeline we adopt. From the left, the monitoring platforms offer raw data composed of quantities and entities. We first pre-process the entities mapping them in a latent space to extract meaningful features in a self-supervised way. We feed the resulting entity embeddings and the original quantities to a multi-modal autoencoder to reduce the dimensionality of the data and the redundant information the different inputs carry. 
The latter includes quantities adaptation modules projecting the raw features to a shared latent space together with the entity embeddings.
We train the MAE in a self-supervised way to produce a compact representation, \ie the multi-modal embeddings. This intermediate representation is then used to solve the final downstream tasks. Note that, in contrast with the common end-to-end training used in most state-of-the-art models, we train the models in each stage of our pipeline independently in a modular way.

In a nutshell, we implement the vision we introduced in~\citep{houidi2022towards}: we move forward toward the adoption of a common DL architecture that is useful for traffic analysis. To demonstrate the feasibility of this approach, we focus on Traffic Classification (TC) uses cases that we select to objectively compare our solution against state-of-the-art alternatives. Results show that our \mae architecture performs better or on par with the specialised models.

\begin{figure*}[!t]
    \centering
    \begin{subfigure}[b]{.22\textwidth}
        \centering
        \includegraphics[width=\linewidth]{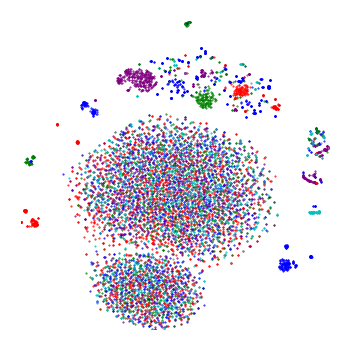}
        \caption{Quantities\\5-NN $p_C=0.32$}
        \label{fig:tsne_quant}
    \end{subfigure}
    \begin{subfigure}[b]{.22\textwidth}
        \centering
        \includegraphics[width=\linewidth]{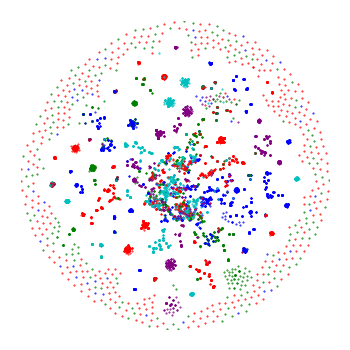}
        \caption{Entities\\5-NN $p_C=0.82$}
        \label{fig:tsne_ent}
    \end{subfigure}
    \begin{subfigure}[b]{.22\textwidth}
        \centering
        \includegraphics[width=\linewidth]{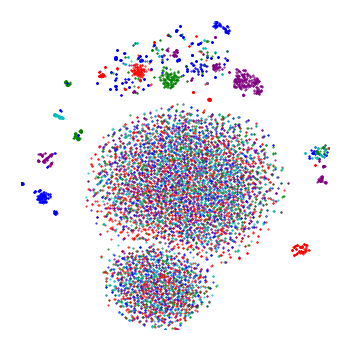}
        \caption{RawCat\\5-NN $p_C=0.32$}
        \label{fig:tsne_raw}
    \end{subfigure}  
    \begin{subfigure}[b]{.22\textwidth}
        \centering
        \includegraphics[width=\linewidth]{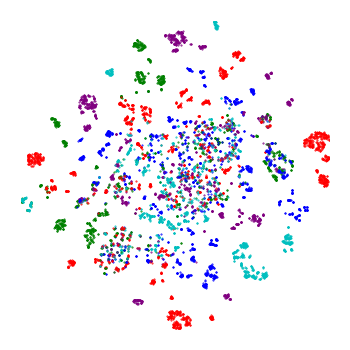}
        \caption{MAE\\5-NN $p_C=0.84$}
        \label{fig:tsne_mae}
    \end{subfigure}  
    
    \caption{t-SNE projection of different representations of the original data. Each dot is a flow. Colours represent the flow class. (a) Using only quantities results in mixing classes in the two large central groups; (b) entities offer better separation of classes, but still mix part of them (surrounding border); (c) a simple concatenation of the two representations introduces noise similar to quantity-only case; (d) \mae improves the representation, offering the best separation among classes.}
    \label{fig:tsne}
\end{figure*}

To give an intuition of how different data representations may be instrumental for, \eg a classic traffic classification task (as used in our validation), \cref{fig:tsne} compares the t-SNE~\citep{vandermaaten2008visualizing} projections using different representations of the original data. Each dot in the figure is a flow, and each colour is a class (details in \cref{s:datasets}). \cref{fig:tsne_quant} and \cref{fig:tsne_ent} show the representation when considering only either quantities or entities separately. For this use case, entities offer better separation of classes than quantities. \cref{fig:tsne_raw} shows what happens when we use a simple concatenation of the two features as typically done in the literature. Confusion appears and classes get mixed again, as one could expect given the high dimensionality of the concatenated data. At last, \cref{fig:tsne_mae} shows the benefits brought by the \mae. In a nutshell, the resulting intermediate representation leverages the information of both quantities and entities, producing a compact representation that already enables a good separation of classes even in such a simplistic t-SNE projection. We will show later that this representation also paves the road for excellent results in specific downstream tasks with other ML algorithms. 

Summarising our contributions:
\begin{enumerate}
    \item We present a simple and streamlined DL architecture that one can use to solve generic traffic analysis-related tasks, avoiding the need to design custom and specific DL architectures for each problem.
    \item This architecture supports both entities and quantities and automatically projects the information they carry into a compact embedding space.
    \item We test our approach in three traffic classification use cases to illustrate the goodness of the embeddings, showing that they carry information that successfully allows one to solve supervised traffic classification tasks.
\end{enumerate}

Finally, we offer to the community all our MAE models, together with all our source code to let other researchers experiment with different tasks and use cases.\footnote{\url{https://github.com/SmartData-Polito/multimodal-ae-for-networking}.}

After presenting related work in \cref{s:soa},
\cref{s:rl} describes the generic building blocks we propose to derive embeddings of typical entities found in traffic traces, and the architecture of the Multi-modal Autoencoder to merge various input data into intermediate embeddings. 
After describing the use cases and datasets used for our validation experiments in \cref{s:datasets}, we validate the proposal with downstream traffic classification tasks in \cref{s:rl_results} and highlight the benefit of the multi-modality compared to individual measurements in \cref{s:ablation}. 
We then discuss the quality of the multi-modal embeddings in \cref{s:properties} and illustrate their applicability in other scenarios (\eg shallow learners) in \cref{s:clustering}. \cref{s:limitations} lists the limitations of the approach and some open questions not addressed in this work. In \cref{s:conclusion} we draw our final considerations.
\newcommand*\fullcirc{\tikz\fill (0,0) circle (1.0ex);} 
\newcommand*\emptycirc{\tikz\draw (0,0) circle (1.0ex);} 

\section{Related Work}
\label{s:soa}

Recent research work focuses on automating traffic analysis through DL models, often borrowing techniques from diverse fields like NLP and CV. While not comprehensive, we report in~\cref{tab:soa} the most notable papers of the last seven years. We focus on top networking conferences and journals, listing only the most relevant papers dealing with supervised traffic classification. 

\begin{table}[t]
\scriptsize
\centering
\caption{Related studies on DL techniques for automatic TC.}
\label{tab:soa}
\begin{tabular}{lc|ccc|cc|cc}
\toprule
& & \multicolumn{3}{c|}{\textbf{Quantities}} & \multicolumn{2}{c|}{\textbf{Entities}} & \multicolumn{2}{c}{\begin{tabular}[c]{@{}c@{}}\textbf{Downstream}\\ \textbf{Classification}\end{tabular}} \\ 
\midrule
& \textbf{Year} & \textbf{Stats.} & \textbf{L7 Payload} & \textbf{Sequences} & \textbf{IP Addr.} & \textbf{Ports} & \textbf{Hosts} & \textbf{Traffic} \\
\midrule
Vu et al.~\citep{vu2017deep}  & 2017 & FE+GAN & -- & FE+GAN  & FE+GAN  &  FE+GAN  &  & \ding{51}   \\
Wang et al.~\citep{wang2017endend}  & 2017 & CNN & CNN & CNN  & 1HE  &  1HE   &  & \ding{51}  \\
Lotfollahi et al.~\citep{lotfollahi2020deep}  & 2017 & CNN/SAE & CNN/SAE & CNN/SAE  & --  &  --   &  & \ding{51}   \\
Hochst et al.~\citep{hochst2017unsupervised}  & 2017 & AE & AE & --  & --  &  --   &  & \ding{51} \\
Gonzalez et al.~\citep{gonzalez2017net2vec} & 2017 & -- & -- & Net2Vec & Net2Vec & --  & \ding{51} &    \\
Ring et al.~\citep{ring2017ip2vec} & 2017 & Word2Vec & -- & -- & Word2Vec & Word2Vec  &   &  \\
Wang et al.~\citep{wang2018hastids}  & 2018 & CNN & -- & CNN+LSTM & --  &  --   &  & \ding{51} \\
Rezaei et al.~\citep{rezaei2019deep}  & 2018 & CNN & CNN & CNN  & --  &  --   &  & \ding{51}  \\
Aceto et al.~\citep{aceto2019mimetic} & 2019 & -- & GRU & CNN & -- & --  &  & \ding{51}  \\
Aceto et al.~\citep{aceto2019mobile} & 2019 & -- & GRU & CNN & -- & --  &  & \ding{51}  \\
Cohen et al.~\citep{cohen2020dante} & 2020 & -- & -- & -- & Word2Vec & Word2Vec  &  &   \\
Holland et al.~\citep{holland2021new}  & 2021 & 1HE  & -- & -- & 1HE  &  1HE   & \ding{51} &  \\
Shahraki et al.~\citep{shahraki2021internet}  & 2021 & CNN  & -- & -- & --  &  1HE+CNN   & \ding{51} &  \\
Aceto et al.~\citep{aceto2019mobile} & 2021 & -- & GRU & CNN & -- & --  &  & \ding{51}  \\
Gioacchini et al.~\citep{gioacchini2021darkvec} & 2021 & -- & -- & -- & Word2Vec & --  & \ding{51} &  \\
Horowicz et al.~\citep{horowicz2022few} & 2022 & CNN & -- & -- & -- & --  &  & \ding{51} \\
Kallitsis et al.~\citep{kallitsis2022detecting} & 2022 & AE & -- & -- & -- & 1HE+AE  & \ding{51} &   \\
Lin et al.~\citep{lin2022etbert} & 2022 & FE & -- & BERT & -- & --  &  & \ding{51}  \\
Gioacchini et al.~\citep{gioacchini2023idarkvec}  & 2023 & -- & -- & -- & Word2Vec & --  & \ding{51} &  \\
Gioacchini et al.~\citep{gioacchini2023exploring} & 2023 & TGNN & -- & TGNN & TGNN & TGNN  & \ding{51} &  \\
Li et al.~\citep{li2023fusiontc} & 2023 & FE & -- & FE & -- & --  &  & \ding{51}  \\
Yang et al.~\citep{yang2023network} & 2023 & AE & -- & GRU & -- & --  &  & \ding{51}  \\
Zhao et al.~\citep{zhao2023yet} & 2023 & -- & BERT & BERT & -- & --  &  & \ding{51}  \\
Huoh et al.~\citep{huoh2023flowbased} & 2023 & GNN/AE & GNN/AE & GNN/AE & -- & --  &  & \ding{51}  \\
Pang et al.~\citep{pang2023multimodal} & 2023 & GNN & BERT/GNN & -- & -- & --  &  & \ding{51}  \\
\midrule
This work & & FC & GRU & CNN & Word2Vec & Word2Vec  & \ding{51} & \ding{51}  \\
\bottomrule
\\
\multicolumn{9}{l}{\multirow{2}{*}{\begin{tabular}[c]{@{}l@{}}$\dagger$ FC: Fully Connected; GRU: Gated Recurrent Unit; AE: Autoencoder; 1HE: One Hot Encoded;\\GAN: Generative Adversarial Network; SAE: Sparse Autoencoder; FE: Feature Engineering;\\TGNN: Temporal Graph Neural Network\end{tabular}}}
\end{tabular}
\end{table}

We observe many works that propose custom and specific DL architectures for host and traffic classification, i.e., labelling traffic either at the host level (\eg IP address classification) or flow level (\eg application identification). They then apply the classifiers on network intrusion detection, OS identification, etc. -- see~\citep{rezaei2019deep,ferrag2020deep} which present generic surveys. 

We split the normally used input features into quantities and entities. As emerges from \cref{tab:soa}, most of the prior work relies only on quantities, ranging from statistics like the flow size, the first $k$ packets sizes, packet inter-arrival times, and generic sequences of such quantities~\citep{aceto2019mobile,lotfollahi2020deep,hochst2017unsupervised,rezaei2020how,aceto2019mimetic}.

The architectures used to manage quantities confirm the increasing trend of adopting DL techniques, often adapting techniques first introduced in research fields different from networking. 
To extract temporal and spatial correlations from \textit{statistics} of packets streams (\eg minimum, average, maximum of a set of observations), many works adopt Convolutional Neural Networks (CNN)~\citep{horowicz2022few,wang2017endend,lotfollahi2020deep,rezaei2019deep} in a CV style. These works select a CNN architecture as an end-to-end classifier.

For \textit{application payload} (L7), Recurrent Neural Networks (RNN), like Gated Recurrent Unit Networks (GRU) are nowadays preferred. These architectures are successfully used in NLP for extracting \emph{character embeddings}~\citep{nguyen2020hierarchical} from the words they appear in. The intuition is that the sequence of bytes in packet payload matters, and RNN/GRU can leverage such information. Similarly, to manage \textit{sequences} of measurements (\eg sequence of the length of first $k$ packets in a flow), Convolutional Neural Networks (and 1D-CNN in particular) are the typical choices.

To learn better representations, some recent works~\citep{kallitsis2022detecting,lotfollahi2020deep,hochst2017unsupervised,huoh2023flowbased} rely on Autoencoders, which allow projecting input features to a more compact latent space discarding uninformative features, or on transformers~\citep{lin2022etbert,zhao2023yet,
pang2023multimodal}, like BERT~\citep{devlin2019bert}.
Since we aim at managing generic input statistics, we will use a simple Fully Connected (FC) layer for the embedding creation, showing it suffices to compress the statistics and remove redundancy in input.

Few researchers started using entities in the classification pipeline (\eg source IP address, or targeted ports). However, they represent such categorical features through One-Hot Encoding (1HE)~\citep{kallitsis2022detecting,wang2017endend,vu2017deep,holland2021new,shahraki2021internet} where each distinct IP address or port is treated as a binary category. Such an encoding technique is affected by scalability problems. Indeed, in the case of IP addresses, the size of the representation might increase up to $2^{32}$ ($2^{128}$) when encoding the whole IPv4 (IPv6) address space. This leads to (i) computational and memory issues and (ii) the \emph{curse of dimensionality} problem~\citep{bellman1966dynamic}. 

Another emerging trend is the adoption of Word2Vec~\citep{mikolov2013efficient,mikolov2013distributed} or Graph Neural Networks (GNN)~\citep{pang2023multimodal,wu2021comprehensive} to produce embeddings for network entities~\citep{ring2017ip2vec,gioacchini2021darkvec,cohen2020dante,gioacchini2023idarkvec,gioacchini2023exploring}. They provide rich representations of entities by solving a self-supervised task (\ie requiring no labels). We here adopt Word2Vec as a solution to learn embeddings from entities. Our MAE architecture is however generic and can encompass other representations too (see our preliminary work on GNN embeddings~\citep{gioacchini2023exploring}). Indeed, GNNs and others have the potential to create novel features that our MAE architecture would seamlessly integrate into the learning pipeline.   

Finally, toward the direction of multi-modality, in which heterogeneous features not originally comparable are merged in a common latent space, only a few works~\citep{li2023fusiontc,yang2023network} started exploring data fusion approaches (see also~\citep{gao2020survey} for a generic survey).
A common drawback of previous efforts is that they propose ad-hoc, end-to-end DL architectures. During the training phase, the classifier learns its internal representation which results in the best separation of the classes in the original feature space. Even though such representations provide often excellent results for the specific task, they are tied to the considered scenarios and cannot be applied to other cases.

The main difference in our work from the previous DL proposals is that we design a \textit{generic DL architecture} to generate \textit{intermediate embeddings} from the original \textit{multi-modal data}. We build on our previous work~\citep{houidi2022towards} but extend and generalise the approach by proposing a multi-modal autoencoder as architecture to derive generic embeddings from raw features. Such embeddings can then be used to solve the downstream task not only with DL approaches but also with simpler shallow learning models. They pave the road for a single generic architecture that could be used for heterogeneous tasks.

\section{Representation Learning for Traffic Analysis}
\label{s:rl}

The prevalent approach in DL-based solutions for network traffic analysis involves crafting ad-hoc architectures customised for specific tasks.
However, these tailored solutions, due to the diverse range of tasks involved, frequently necessitate extracting specific features through a time-consuming feature engineering process.

In contrast, we advocate the usage of representation learning. We propose a generic and flexible architecture to produce an intermediate compact representation of network data by merging the informative content of heterogeneous features (\ie entities and quantities). This intermediate representation serves the downstream task, reducing the effort required in the feature engineering stage and paving the road for unexplored solutions involving both quantities and entities. 

The choice of Autoencoders as the key model for the proposed architecture is straightforward due to their self-supervised learning abilities and capacity for dimensionality reduction.
They efficiently compress input features, retaining crucial information and minimising the training cost required by more complex architectures, such as transformers. 

\begin{figure}[!t]
    \centering
    \includegraphics[width=\linewidth]{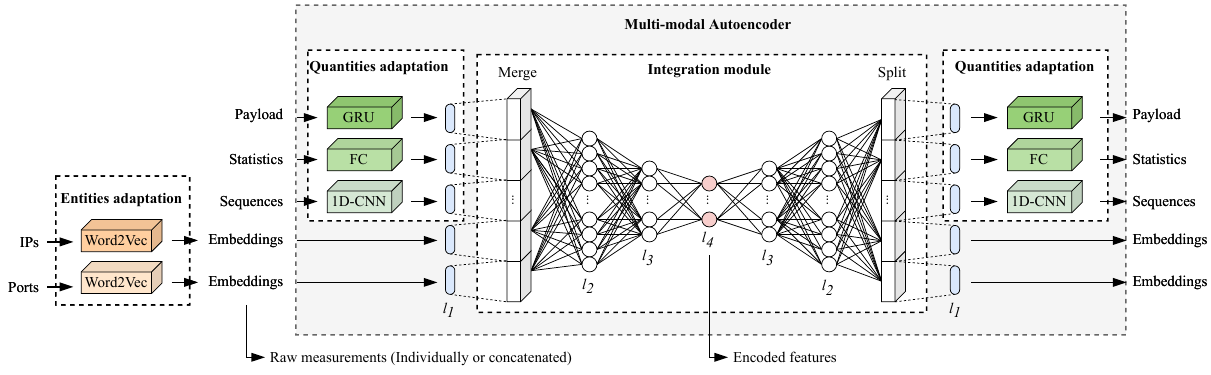}
    \caption{Multi-modal autoencoder architecture.}
    \label{fig:mae}
\end{figure}

Our representation learning architecture is logically composed of two main blocks as sketched in~\cref{fig:mae}. First, we consider \textit{adaptation modules} whose goal is to extract useful representations that are tailored to each input type, where a specific DL architecture deals with each specific traffic data. Each module will produce as output a representation of the original data in a space of $l_1$ dimensions.
The second step consists of an \textit{integration module} that merges and compresses the information captured by each adaptation module into a compact embedding space.

The overall architecture results in a Multi-modal Autoencoder, which is self-trained to minimise the reconstruction error between the input measurement samples (on the left) and the reconstructed output samples (on the right). Once trained, we discard the decoder and use the encoder to compute the embeddings, mapping each input sample into the intermediate representation space in $\mathbb{R}^{l_4}$ (coloured nodes in~\cref{fig:mae}). We call this representation $E(s)$.

\subsection{Raw measurements}
We assume a set of monitors $i\in I$ gathers raw measurements $M^i(s)$ concerning \textit{samples} $s\in S$. 
Samples $s$ are the objects of the final downstream task.
Samples can be TCP/UDP flows, hosts (identified by their IP address), domain names, etc. 
The $i$-th module can be a DPI monitor that extracts per packet information (\eg packet length, source/destination IP addresses, protocol fields, etc.), a flow monitor, such as Netflow that aggregates \textit{statistics} about the flow properties (\eg flow duration, flow size, client/server IP addresses, mean packet size, etc.), or a firewall that observes the \textit{sequences} of packets reaching the internal networks, \eg sequences of ports contacted by a client, sequences of IP addresses reaching a web server, etc.

In general, the $i$-th measurement module will expose a set of measurements for each sample $s$:
$$M^i(s)=\{m^i_1(s),m^i_2(s),\ldots,m^i_{n_i}(s)\}\in \mathbb{R}^{n_i},$$
where $n_i$ is the number of measurements such module extracts.

\cref{tab:measurements} exemplifies some measurements that our architecture supports. 
As previously advocated, network data entails two distinct modalities, \ie entities and quantities.  
Quantities are real-valued numbers that convey some salient characteristics.
Quantities are already in numerical form and suitable for ingestion into ML models.
Some classic quantities include statistics at the flow level like minimum, maximum, and average packet size, or sequences of numerical values like the length of the first $k$ packets in a flow.

Conversely, entities represent categorical features that are not in an ordered relationship. They are traditionally represented in ML tasks using simple One-Hot Encoding. Such a representation fails, however, to capture important properties of the entities. For example, IP addresses or TCP/UDP ports are typical entities used for several traffic classification problems. However, One-Hot Encoding does not scale.
Noticing the similarity between entities and words in natural language, we promote the use of NLP model pre-training to learn useful representations from sequences of entities, hence the need for some pre-adaptation layer (see below).

For entities, we consider \textit{sequences of observations} of samples, let them be IP addresses, ports,  DNS names, or others. The order and context in which such entities co-occur carry valuable information worth feeding as input to the ML model to increase its chances of succeeding in the task. For instance, a sequence of server names contacted over time by a client may reveal the application such client is running. Similarly, the sequence of ports a sender checks on a server may reveal which type of scan it performs.

\begin{table}[t]
    \scriptsize
    \centering
    \caption{Examples of typical features exposed by network monitors that we can consider to create an embedding of samples.}
\begin{tabular}{lll}
\toprule
 & \textbf{Measurement} & \textbf{Description} \\
 \midrule
\multirow{10}{*}{\textbf{Quantities}} & Payload & Raw values of first n bytes of the L7 payload \\ \cmidrule{2-3}
 & \multirow{4}{*}{Statistics}& Min, max, avg, std of packet size per flow \\
 &  & Min, max, avg, std of packet inter-arrival time per flow \\
 &  & \begin{tabular}[c]{@{}l@{}}Min, max, avg, std of ports contacted by clients \end{tabular} \\
 &  & Min, max, avg, std, ... of other metrics \\
 \cmidrule{2-3}
 & \multirow{4}{*}{Sequences} & Length of the first $k$ packets per flow \\
 &  & Inter-arrival time of the first $k$ packets per flow \\
 &  & TCP window size of the first $k$ packets per flow\\
 &  & Other quantities over time \\

\midrule

\multirow{6}{*}{\textbf{Entities}} & IP address & Sequence of client (server) IP addresses  observed over time \\ \cmidrule{2-3}
 & Ports & Sequence of client (server) ports observed over time \\ \cmidrule{2-3}
 & AS & Sequence of DNS names resolved over time \\
  \cmidrule{2-3}
 & Other & Sequence of other entities over time \\
 \bottomrule
\end{tabular}
\label{tab:measurements}
\end{table}

\subsection{Adaptation modules}
\label{sec:adaptation}
We propose to use specifically designed DL adaptation modules to project each feature into a common space of size $l_1$ before integrating them using the MAE. In the following, we describe this specific step for each input type. For a more detailed background and formalisation of the architecture used in the adaptation modules refer to \ref{appx:adaptation}.

\subsubsection{Sequences of Entities adaptation}

We leverage the fact that entities are typically observed as a \textit{sequence} over time, \eg the sequence of server IP addresses contacted by a client or the sequence of ports scanned during a port-scan attack.
We here advocate the use of language model pre-training to learn representations for each entity (\eg the IP address or the port) by exploiting their co-occurrences over time as observed in \textit{historical data}. In particular, we leverage Word2Vec~\citep{mikolov2013efficient,mikolov2013distributed}, an NLP technique relying on artificial neural networks to map words occurring in a sentence into a \emph{context-aware} latent space (\ie words belonging to similar contexts appear close in the latent space) in a self-supervised manner.\footnote{Note that other representations, e.g., obtained with GNNs~\citep{gioacchini2023exploring}, could be used in this step. Including new adaptation modules in our architecture is trivial and we leave the benchmark of their impact for future work.} By defining the context of a target word as the set of the $n$ words surrounding it in a sentence, we adopt the \emph{skip-gram}~\citep{mikolov2013efficient,mikolov2013distributed} technique, which consists of predicting the probability that a word belongs to the context of the target word. $n$ is a user-defined parameter impacting the granularity of the detected context.

Leveraging Word2Vec, given a sequence $\{s_1,s_2,s_3,\ldots \}$ of entities observations, we map each entity $s_j \to u_j\in \mathbb{R}^V$ where $u_j$ is the embedding of $s_j$ in the $V$ dimension space.
As said, prior works have shown that Word2Vec can extract a valuable representation from sequences of IP addresses~\citep{ring2017ip2vec,gioacchini2021darkvec,gioacchini2023idarkvec,kallitsis2022detecting,cohen2020dante}.
Here, to represent sender IP addresses, we consider their co-occurrence when targeting the same port or host in a given time window.

Generalising the approach to all entities, we propose to use it systematically to assign a compact yet rich feature vector to each entity. For instance, we generate port embeddings feeding Word2Vec with the sequence of server ports targeted by each sender as proposed by authors of~\citep{cohen2020dante}. 

The dimension $l_1$ of the space is a parameter that impacts the embedding quality. Any choice of $l_1\in[50-200]$ results sufficient to produce valuable embeddings. We investigate its impact in~\cref{fig:gridsearch}. 

Given that Word2Vec is a self-trained architecture, we train the embeddings separately from the MAE and give the results as input directly to the integration module.

\subsubsection{Quantities adaptation}
Quantities adaptation modules are neural networks tailored to each quantity type and trained with self-supervision to extract salient features from quantities.

\paragraph{Application payload adaptation}

Even with the increasing adoption of encrypted protocols (such as QUIC, H3, TLS 1.3, DoH, etc.), network traffic still includes portions that are transported in plain text, such as specific protocol headers. Indeed, previous works~\citep{aceto2019mobile,nascita2021xai} have shown that training a NN directly with raw payload results in models able to perform \eg flow classification with high accuracy. Our architecture aims to streamline the traffic analysis problem by combining multiple types of features. We thus introduce an adaptation module that handles payload directly, leaving to the Integration Module (described next) the task of learning the salient information from such features (if any). 

Given a flow and the first $n$ bytes of payload, we propose adapting it using an embedding layer, which allows mapping the categorical values taken by each byte into a compact space, followed by an RNN, which allows leveraging of information carried by consecutive portions of the payload, as proposed in ~\citep{aceto2019mobile,nascita2021xai}.
Here, we use RNNs as they are widely used for learning sequential data prediction problems, but any sequence-friendly neural network architecture could fit the purpose.

In our experiments, we use a GRU which is like a Long Short-Term Memory (LSTM) with a forget gate that allows the network to forget very old correlations. It has fewer parameters than an LSTM and often converges faster. The essential operation of LSTM or GRU networks can be considered to hold the required information and discard the information that is not helpful for the prediction. 
The intuition here is to let the GRU exploit the correlation found in nearby bytes in the payload, \eg server names found in clear in the Server Name Indicator (SNI) in TLS, or specific values in TCP Options or IP headers.  

\paragraph{Traffic statistics adaptation}
Consider statistics $M^i(s)=\{m^i_j(s)\}$ extracted by the monitor $i$ for a sample $s$. Examples of $m^i_j(s)$ include the minimum, maximum, average, standard deviation, histograms, etc. For instance, if $s$ is a TCP or UDP flow, the monitor computes such statistics on all flow packets. Instead, if $s$ is a host, statistics derive from the set of contacted ports, the set of contacted server IP addresses, etc.

Given a set of monitors $I$, we expect such statistics to be correlated among them (\eg minimum, mean, median, etc.), and among different measurements (\eg statistics on the number of packets and the number of bytes, on sent and received traffic, etc.). As such, we encode such statistics using a Fully Connected (FC) layer to reduce such correlation while compressing and adapting the input data to the integration module. For this, as input, we feed the FC with the concatenation of all statistics for a given sample $s$, \ie $M^I=\{M^1(s),M^2(s),\ldots\}$. As output, we obtain a projection in a reduced space of size $l_1$.

\paragraph{Sequence of quantities adaptation}
When the order of measurements is important, it is common to use sequences of observations of a sample $s$ as input features. For instance, the sequence of the first $k$ packet lengths and inter-arrival times of a given flow is often used for the flow classification~\citep{aceto2019mobile,wang2018hastids,rezaei2020how}.

As commonly done in past works (see~\cref{tab:soa}), we encode such sequences using a CNN. We use a one-dimensional CNN (1D-CNN) that lets us obtain compact information from such correlated measurements and reduce the overall complexity of the model in terms of the number of parameters.
As before, the input of the CNN will be the sequences of all measurements. The output will be a $l_1$ projection that will be given as one of the inputs to the integration module.

\subsection{Integration module}

The goal of the integration module is to fuse different measurements that derive from different monitoring modules and their corresponding adaptation layer. This is particularly important when the number of input measurements is large, \ie when commonly done when using DL approaches that use all possible data the various monitors produce without performing a prior feature selection or engineering.

To this goal, we propose to use a symmetric autoencoder architecture made of five layers, each of $l_2, l_3, l_4, l_3, l_2$ neurons with $l_2 > l_3 > l_4$. As input, it takes the output of each adaptation module. The autoencoder then compresses this information so that the inner layer of $l_4$ neurons produces a compact representation of the input measurements $E(s)$. To train such a network, the architecture is symmetric so that, starting from the encoded features, a symmetric structure rebuilds the input structure. 
The training of the autoencoder is self-supervised and does not require any labels.

The overall system results in a multi-modal autoencoder that includes the quantities adaptation modules as well. Except for the entities adaptation modules which are pre-trained separately, the training of the MAE includes both the quantities adaptation modules and the integration module. The overall system is trained to minimise the reconstruction error from the original measurements to the reconstructed outputs. 

Since the \mae is an Autoencoder, whose goal is reconstructing the input data at the output layer, we essentially solve a regression task. Hence, we use a linear activation function in all the decoding output layers (\ie quantities adaptation outputs and entity embeddings) to achieve an unbounded range of output values favouring the input reconstruction. Additionally, we use the Mean Squared Error (MSE) as the loss function of each adaptation module. 
To balance the contribution of each measurement (quantities and embedded entities), we weigh the losses proportionally to the number of features (embedding size) obtaining the final weighted MSE for the whole \mae.

Notice that each adaptation module is designed specifically for the raw measurement to adapt. For a detailed description of the adaptation modules, refer to \ref{appx:architectures}. In the following, we compare the performance of using different feature sets when facing three downstream traffic classification tasks.
\section{Validation tasks and datasets}
\label{s:datasets}

\begin{table*}[t]
\footnotesize
    \centering
    \caption{Downstream tasks used in our validation and summary of the datasets.}
\begin{tabular}{lcccccc}
\toprule
\textbf{Task} & 
\textbf{Dataset} & \textbf{Sample} & \textbf{Label} & \textbf{\# Samples} & \textbf{\# Classes} & \textbf{Balance} \\
\midrule
$T_1$ & \texttt{MIRAGE}~\citep{aceto2019mirage} & Flows & Mobile apps & 44k & 16 & 0.94 \\
$T_2$ & \texttt{DARKNET}~\citep{gioacchini2021darkvec} & IP addr. & Coord. groups & 14k & 13 + 1 & 0.4 \\
$T_3$ & \texttt{ISCX}~\citep{noauthornodatevpn} & Flows & Traffic type & 609 & 5 & 0.82 \\
 \bottomrule
\end{tabular}
\label{tab:dataset}
\end{table*}

We here focus our evaluation on supervised traffic classification tasks. These tasks allow us to compare the performance of our architecture against state-of-the-art solutions objectively. Our architecture is however generic and can be employed in other downstream tasks too.     

To create a solid baseline, we select a set of tasks from previous articles that proposed end-to-end classifiers and contribute open datasets to the community.
We select three traffic classification tasks requiring different sets of features, whose datasets are well-documented.\footnote{Ideally, the power of alternative representations could be better compared using a single dataset in multiple downstream tasks. However, the networking community still misses established and generic datasets and benchmarks.}   
The presence of labels (\ie a ground truth) allows us to objectively evaluate each approach, assessing and comparing the quality of the representations used as input.

\cref{tab:dataset} summarises the downstream tasks and datasets we use. It shows the task number, dataset name and reference, the target samples to be classified, and the type of labels, followed by some statistics about the dataset size, the number of classes, and the balancing coefficient.\footnote{The coefficient is defined as the Shannon entropy (with respect to the classes of the problem) divided by the logarithm of the number of classes. The coefficient is closer to 1 when classes are balanced.}

We select cases with different numbers of classes, class balances, and dataset sizes. Next, we briefly describe the tasks and provide more details about the datasets.   

\subsection{$T_1$: Identification of mobile apps traffic}

The identification of the applications generating traffic is one of the classic examples of TC. This problem has been approached with multiple formulations, but in general, the goal is to identify which application generated particular \textit{traffic flows}.

We use as our first task the problem and dataset presented in~\citep{aceto2019mirage}. The authors focus on classifying TCP and UDP/QUIC flows as generated by one among multiple \textit{mobile apps}. They collected two years of traffic from volunteers who installed a custom monitoring app, from 2017 to 2019, releasing the \texttt{MIRAGE-2019} dataset (hereafter called \texttt{MIRAGE}).
Each flow in the dataset is labelled with the name of the mobile app that generated it or as background traffic (filtered out in a post-processing step). There are 16 different applications, from social network apps to e-commerce and news apps.

The dataset comprises 44k flows (our sample set $S$), with well-balanced classes  (coefficient equal to 0.94). Each flow is characterised by the sequence of packet inter-arrival times and length, the sequence of the TCP receiver window values, the application payload of up to 32 bytes, the server IP address, and its /24 subnet.

\subsection{$T_2$: Classification of darknet traffic}

Our second task is the classification of darknet traffic~\citep{gioacchini2021darkvec,kallitsis2021zooming,kallitsis2022detecting}. Darknets are networks used to monitor \textit{unsolicited traffic}, such as Internet scans and attacks. They are composed of IP addresses announced in routing protocols, but without hosting services. The identification of interesting events in darknet traffic is hard, as hundreds of thousands of attackers and scanners continuously reach the darknet IP range. However, previous work has shown that IP addresses belonging to a limited number of botnets and scanners dominate this traffic. 

The task is to classify the sender IP addresses (our sample set $S$) as belonging to one of the multiple classes of known scanners and attackers.
For this problem, authors have leveraged different measurements, including raw per-sender traffic statistics and features obtained using embedding techniques such as Word2Vec~\citep{gioacchini2021darkvec} or feature engineering and autoencoders~\citep{kallitsis2021zooming,kallitsis2022detecting}.

We here rely on the public dataset~\citep{gioacchini2021darkvec}, which we call \texttt{DARKNET}. There are 13 categories of known senders which include several benign security services, research scanners, and known botnets. The dataset also includes one extra class with all \textit{unknown} IP addresses. 

In total, \texttt{DARKNET} collects one month of data reaching one /24 darknet. By considering all senders that sent at least five packets in such period, we have 14k sender IP addresses, among which there are 5k addresses from known classes; the rest belong to the unknown class. In contrast to the other tasks, this dataset shows strong class unbalance, with a coefficient equal to 0.4.

As measurements, $M^i(s)$, here we consider monitors that expose the sequence of sender IP addresses as they appear in the trace, statistics of the TCP/UDP ports they target, statistics of packet length each sender sent, and the /24 subnet the sender belongs to.

\subsection{$T_3$: Classification of traffic flows}

Our last task deals with classifying traffic flows (our sample set $S$), \ie video streaming, voice, browsing, etc. This task is also a classic in TC and is particularly hard because most services rely on encrypted protocols (\eg TLS).
Multiple authors have proposed methodologies, usually based on supervised classifiers that take packet-level measurements as input. 

We here rely on the \texttt{ISCXVPN2016}~\citep{noauthornodatevpn} (\texttt{ISCX} for brevity) dataset introduced in~\citep{horowicz2022few,noauthornodatevpn}.
We consider only the non-tunnelled traffic in the dataset, which includes five traffic categories: VOIP, streaming, chat, mail, and file transfer.
The dataset is very limited, with 609 flows that are well-balanced among the five classes (coefficient 0.82). Such a small dataset poses a challenge for DL training, and we thus use it to check how the MAE behaves with scarce data. 

Here, each TCP or UDP flow is labelled with a category. Each flow $s$ is characterised by several measurements $M^i(s)$: the server IP address and its /24 subnet, the first 32 bytes of application payload, statistics collected from TCP payload, and the sequence of the first $k=32$ receiver window size, packet inter-arrival times and length.

\section{Assessing the MAE Embeddings}
\label{s:rl_results}

\subsection{Experimental Setup}

Using the three supervised downstream tasks, we evaluate the results obtained with the \mae embeddings and compare them against traditional approaches. In this section, we provide a high-level summary of the results using our approach and state-of-the-art alternatives. In the next section, we present an ablation study to assess the impact of the different components of our method.

Given a set of input measurements $M(s)$, we train a classification model 
$f: f(s) \to C(s)$, being $C(s)$ the class of $s$.  For each task, we present two experiments: 
(i) We naively achieve a multi-modal representation of the different inputs by simply concatenating the embedded entities together with remaining raw data to obtain a single input vector $M(s)=concat(M^1(s),M^2(s),\ldots)$. Then, we train a model $f: f(M(s)) \to C(s)$; 
(ii) we consider the representations $E(s)$ obtained with our \mae. Here, we train a model $f: f(E(s)) \to C(s)$.

For each $M(s)$, we run 5-fold cross-validation experiments reporting average statistics. When validating models for $T_2$, we ignore the \textit{unknown} class as done by the original authors of the datasets.
For $T_1$ and $T_3$, we use stratified sampling to guarantee that flows belonging to the same \textit{application session} appear either in the training or in testing sets.

To achieve an architecture that avoids excessive computational times and memory requirements, we fix the sizes of the hidden layers of the integration module $l_2$ and $l_3$ to 512 and 256 neurons, respectively. These numbers have been defined after a coarse grid search discussed in~\ref{s:complexity}.

After training the entities adaptation modules and the \mae, we extract the resulting embeddings of a sample $E(s)$ and we use it as input to the DL classifier $f$.
We consider the same DL architecture for $f$ in all the experiments.
Specifically, we use a fully connected feed-forward network with two hidden layers (of 512 and 256 neurons, respectively), and an output layer with several neurons equal to the number of task classes. To prevent overfitting we use a dropout of 30\% for each layer. The activation function of the hidden layers is a standard ReLU, whereas the output layer is activated through the Softmax activation function. We use class balancing during training to weight the loss function and minimise the categorical cross-entropy using an ordinary Adam optimiser~\citep{kingma2017adam}.

\subsection{Classification performance}
\label{ss:downstream}

\begin{table}[!t]
\centering
\footnotesize
\caption{Macro and average F1-Scores across tasks. \textit{Reference} is the macro average F1-Score reported in the original work. For $T_3$, this is missing, since we use only a fraction of the original dataset. The best results are in \textbf{bold}.}
\begin{tabular}{lc|cc|cc|cc}
\toprule
 & \textbf{} & \multicolumn{2}{c|}{\textbf{Macro F1-Score}} & \multicolumn{2}{c|}{\textbf{Weighted F1-Score}} & \multicolumn{2}{c}{\textbf{Trainables}} \\
 & \textbf{Reference} & \textbf{Concat.} & \textbf{MAE} & \textbf{Concat.} & \textbf{MAE} & \textbf{Concat.} & \textbf{MAE} \\
\midrule
$T_1$ & 0.88 & \textbf{0.89} & 0.87 & \textbf{0.90} & 0.88 & $29\times 10^4$ & $17\times 10^4$ \\
$T_2$ & 0.96 & \textbf{0.98} & \textbf{0.98} & 0.99 & \textbf{1.00} & $33\times 10^4$ & $17\times 10^4$ \\
$T_3$ & -- & 0.66 & \textbf{0.68} & 0.74 & \textbf{0.75} & $28\times 10^4$ & $17\times 10^4$ \\
\bottomrule
\end{tabular}
\label{tab:results}
\end{table}

In \cref{tab:results} we report the macro and weighted average F1-Scores\footnote{Macro average is the arithmetic mean of the individual class scores, while weighted average includes the individual class sample sizes. The latter better represents results in practice. However, for highly unbalanced cases, it provides biased results for the minority classes.} resulting from the classification tasks, along with the number of trainable parameters of the classifiers as a proxy of the computational cost of classifier training. 

\paragraph*{$T_1$ -- \texttt{MIRAGE} dataset}

Focus on the \texttt{MIRAGE} dataset ($T_1$).
The best results are obtained when using concatenation in a single input vector, reaching 0.89 macro average F1-Score. Yet, the classification using the \mae is very similar to the raw concatenation case, losing 0.02\% of the macro average F1-Score. Intuitively, the compression performed by the \mae reduces the redundancy of information carried by correlated features with a minimal impact on the final results.

The slightly better results of the raw concatenation come at the expense of downstream model complexity. Indeed, the concatenated input layer has about 300 neurons resulting in $29\times10^4$ trainable parameters, while the \mae with just $l_4=64$ input neurons halves the number of trainable parameters down to $17\times10^4$ achieving comparable classification performance.

These results are also in line with the original paper, where authors report 88\% macro average F1-Scores~\citep{aceto2019mimetic} using an end-to-end DL pipeline that relies on the payload and the sequences of packet measurements as input.
The \mae features achieve performance in the same range with a more compressed representation.

\paragraph*{$T_2$ -- \texttt{DARKNET} dataset}

Move now to the \texttt{DARKNET} dataset ($T_2$). Firstly, both the naive multi-modal solution and the proposed \mae result in a slight improvement compared to the baseline. 

The concatenation of all raw measurements here performs slightly worse than using the \mae (weighted average F1-Score) -- specifically, one class gets much worse results.
This happens because some measurements bring unimportant or redundant information for this task, polluting results for the concatenation case. 
This is an expected result given the importance, for ML practitioners, of removing noisy features that may degrade the performance of classifiers. Notice that, along with the slightly improved performance, the \mae allows also a reduction of the number of trainable parameters of $>50\%$ -- $17\times10^4$ compared to the $33\times10^4$ of the raw concatenation.

\paragraph*{$T_3$ -- \texttt{ISCX} dataset}

Finally, check the results obtained with the \texttt{ISCX} dataset ($T_3$). Recall that we have a limited dataset of about 600 samples and thus we expect this case to be a challenging scenario for any machine learning algorithm, in particular those based on DL models. 

Even though the overall classification performance cannot be considered satisfactory (average F1-Score $<80\%$), we can still confirm the considerations drawn from the previous tasks:
The \mae embeddings allow the DL classifier to achieve performance better than the one obtained with the crude concatenation of raw features reducing the number of training parameters from $28\times10^4$ down to $17\times10^4$.

In a nutshell, also in this case the, \mae embeddings capture key characteristics of the data, even in such a small sample, reducing the trainable parameters by $\approx 50\%$.

\vspace{3mm}
\textbf{Takeaway}: The experiments show that for TC problems the same generic \mae architecture can be used to face very different supervised tasks. On one hand, the adaptation modules allow considering as input heterogeneous measurements. On the other hand, the integration module allows compressing the input data reducing downstream model trainable parameters by $\approx50\%$, without losing performance. 
\section{Ablation Study}
\label{s:ablation}

We now delve into an ablation study. Provided that the \mae key advantage is its ability to combine multiple features, coming from the multiple modules, we focus on assessing the impact of adding the integration module. We here consider only the best parameters for the \mae architecture, which have been found via a grid search, summarized in \ref{s:complexity}.\footnote{Another valid ablation study would be the evaluation of the importance of individual layers of a pre-trained \mae on results. We leave this study for future work.}

For each downstream task, we compare the classification performance of \mae embeddings against the performance obtained using a single measurement. These experiments show in particular the impact of our multi-modal strategy on performance. We consider separate classifiers $f^i$ trained using only individual measurements, \ie given $M^i(s)$, we train a model $f^i: f^i(M^i(s)) \to C(s)$. We embed entities using the Word2Vec, and we feed other measurements directly as input to each separate classifier.

\begin{figure}[!t]
    \centering
    \begin{subfigure}[b]{.4\textwidth}
        \centering
        \includegraphics[width=\linewidth]{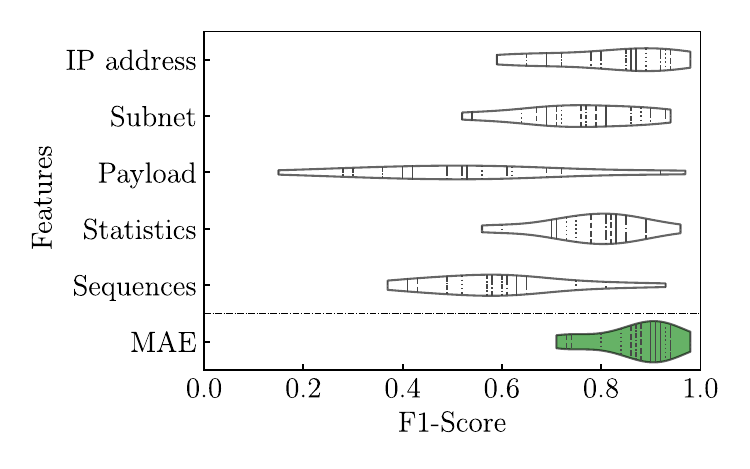}
        \caption{$T_1$, \texttt{MIRAGE} dataset.}
        \label{fig:task1_violin}
    \end{subfigure}
    \begin{subfigure}[b]{.4\textwidth}
        \centering
        \includegraphics[width=\linewidth]{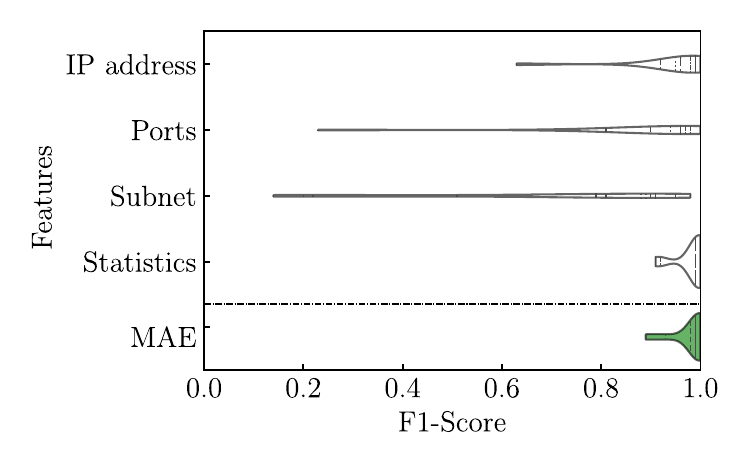}
        \caption{$T_2$, \texttt{DARKNET} dataset.}
        \label{fig:task2_violin}
    \end{subfigure}
    \begin{subfigure}[b]{.4\textwidth}
        \centering
        \includegraphics[width=\linewidth]{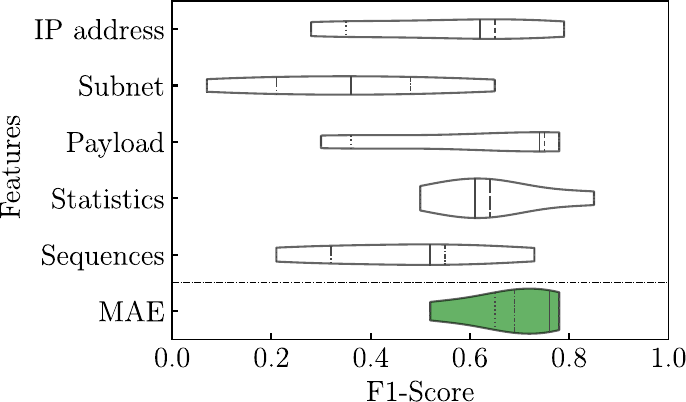}
        \caption{$T_3$, \texttt{ICSX} dataset.}
        \label{fig:task3_violin}
    \end{subfigure}    
    \caption{F1-Score distributions per class for each classification task and different input measurements}
    \label{fig:violinplots}
\end{figure}

\begin{table}[t]
\centering
\footnotesize
\caption{Macro average F1-Scores across tasks when using independent features and multi-modal features. The best results are in \textbf{bold}.}
\begin{tabular}{llccc}
\toprule
 &  & $T_1$ & $T_2$ & $T_3$ \\
\midrule
\multirow{3}{*}{\textbf{Entities}} & \textbf{IP address} & 0.82 & 0.96 & 0.47 \\
 & \textbf{Ports} & -- & 0.91 & -- \\
 & \textbf{Subnet} & 0.76 & 0.70 & 0.54 \\
\midrule
\multirow{3}{*}{\textbf{Quantities}} & \textbf{Payload} & 0.53 & -- & 0.35 \\
 & \textbf{Statistics} & 0.78 & \textbf{0.98} & 0.65 \\
 & \textbf{Sequences} & 0.59 & -- & 0.59 \\
\midrule
\textbf{Multi-modal} & \textbf{MAE} & \textbf{0.87} & \textbf{0.98} & \textbf{0.68} \\
\bottomrule
\end{tabular}
\end{table}

We summarise the results with the three downstream tasks. 
\cref{fig:violinplots} shows a summary of the results using violin plots, whereas \cref{tab:results} reports the details with macro average F1-Scores.  
Each violin in \cref{fig:violinplots} represents a distribution of the F1-Scores for all classes in a given classification task. Empty violins represent the distributions for models $f^i$ trained with separate sets of measurements. Green violins refer to the results of using the \mae as input to the DL classifier. 

\paragraph*{\texttt{MIRAGE} dataset}

Start from the \texttt{MIRAGE} dataset ($T_1$) in \cref{fig:task1_violin}.
When using a single set of raw measurements (top white violins), the distribution of the F1-Score across classes is widely spread.

First, the classifier trained with \textit{entity} (IP) embeddings alone, a feature not used in the original task~\citep{aceto2019mimetic}, achieves better performance than the ones trained with \textit{quantities}, reaching 0.82 Macro average F1-Score. This further motivates our call for the systematic inclusion of entity information in addition to classic quantities.
For quantities, the model trained with flow traffic statistics obtains satisfactory results (0.78 Macro average F1-Scores). 

\mae leads to the best results of 0.87 macro average F1-Score compared to the independent measurements.
This result confirms that, for this use case, the multi-modal architecture with all adaptation modules produces a rich and compact representation, and improves performance over a DL classifier trained on individual measurements.

\paragraph*{\texttt{DARKNET} dataset}

Consider now the \texttt{DARKNET} ($T_2$) dataset in \cref{fig:task2_violin} and \cref{tab:results}. Here, we see again that classifiers trained with separate sets of measurements deliver performance that varies greatly according to the employed measurement set. Only traffic statistics let the classifier reach solid results (macro average F1-Scores of 98\%).

The \mae (with all its adaptation modules) shows an important advantage in this case: The methodology transparently overcomes the presence of uninformative features in these input sets. We see in the green violin plot in \cref{fig:task2_violin} that the distribution of F1-Scores of a DL classifier with \mae encoded features is more compact, achieving excellent results. More important than searching for performance improvements in this particular downstream task, we stress the fact that the \mae embeddings are obtained with a generic feature extraction approach, thus streamlining the procedure for different problems.

\paragraph*{\texttt{ISCX} dataset}

Finally, focus on \texttt{ISCX} dataset ($T_3$) results. We see in \cref{fig:task3_violin} that the various distributions of F1-Scores are much wider than in previous tasks. Contrary to the previous tasks, IP address embeddings have worse performance compared to quantity features (0.46 of macro F1-Score). 

Despite the overall \mae classification performance that remains unsatisfactory (0.68 of F1-Score), also in this case our proposed architecture can capture key characteristics of the data automatically. Indeed, the \mae preserves the informative content of heterogeneous features, discarding useless information and bringing an improvement compared to the individual measurements (from +3\% compared to traffic statistics to +33\% compared to L7 payload).

\vspace{3mm}
\textbf{Takeaway}: The obtained results underscore the advantage of \mae in preserving informative content across diverse features, and in effectively discarding irrelevant or uninformative content. 
The combination of several features results in better performance overall, even in the presence of uninformative features. This is possible thanks to the Integration Module that compresses information discarding the eventual unnecessary data. The complete \mae thus leads to a more streamlined and enhanced representation of network data across various tasks.

\section{Embedding properties}
\label{s:properties}

The previous sections show that \mae embeddings are informative for traffic classification problems. These results raise the question of whether they carry sufficient discriminative power to support other downstream tasks too. Here we test whether the embeddings would support distance-based algorithms. Recall from \cref{fig:tsne_mae} that we are searching for \textit{rich} representations that enable several downstream tasks. To gauge the quality of the embeddings, we investigate the local neighbourhood of samples of known classes, assessing whether their neighbourhood includes samples of the same class.

\subsection{Methodology}

We study how samples \textit{with ground-truth labels} are positioned in the latent space. We define a metric that quantifies the \textit{pureness} of the neighbouring samples as follows. Given a sample $s_i$ of class $c_i$ and the set of its $K$-nearest-neighbours $\mathcal{N}^{K}_i$, we define the $K$-NN class probability for the sample $s_i$ as: 
$$
p_{i,C}(K) \; = \; P( C(s_j)=c_i \; | \; s_j \in \mathcal{N}^{K}_i)
$$
In words, the metric indicates how pure the neighbourhood of $s_i$ is, with $p_{i,C}=1$ indicating that all $K$ neighbours of $s_i$ are of the class $c_i$. 

Given $K$, we compute the macro average of the $p_C(K)$ by computing first the average among all samples of a given class and then averaging over classes. We use the cosine distance to find the neighbours, a metric often used as a distance when dealing with embeddings~\citep{brown2020language,mikolov2013distributed}.

We compare $p_C(K)$ considering (i) the space defined by the concatenation of all raw measurements, (ii) the \mae latent space. 

\begin{figure}[t]
    \centering
    \includegraphics[width=.6\linewidth]{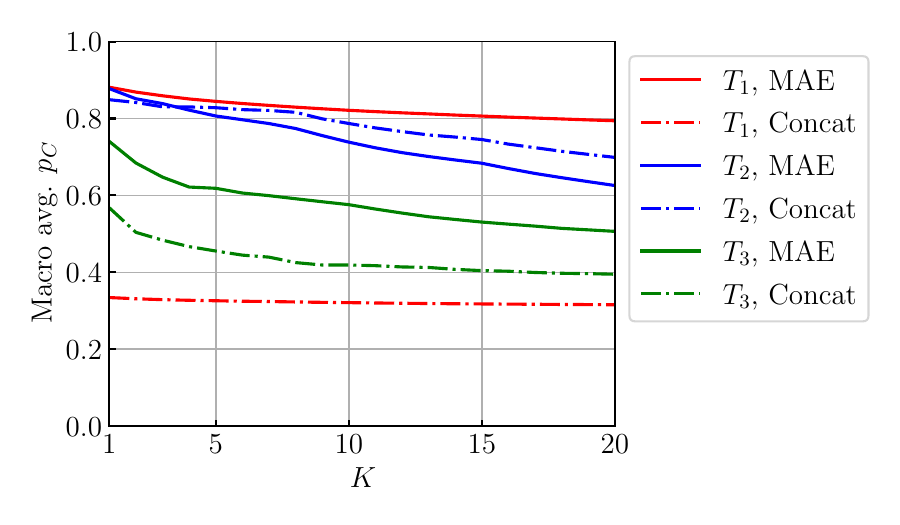}
    \caption{Macro average $K$-NN class probability.}
    \label{fig:knn}
\end{figure}

\subsection{Neighbourhood pureness}

\cref{fig:knn} shows the macro averages of $p_{i,C}$ when varying $K$ for the \mae embedding (solid lines) and the concatenation of raw measurements (dashed lines). Colours distinguish the three use cases. 

Focusing on $T_1$, we see that the neighbourhood of samples is very polluted when using the brute concatenation of raw measurements (red-dashed lines). 
Back to~\cref{fig:tsne_raw}, the {t-SNE} shows some areas with clearly pure clusters but also two large clouds of samples of quite mixed classes. There, we expect distance-based algorithms to likely group elements of different classes in single clusters.

The \mae produces a latent space that completely reorganises the samples, in this case placing elements of the same class nearby in space (see red-solid lines), although it was not trained for this purpose. This is precisely the effect observed in \cref{fig:tsne_mae}. The values of $p_C(5)$ in~\cref{fig:tsne} reflects the improvement. 
Here we expect the \mae embedding to present a clear advantage when applying distance-based algorithms, or even shallow learners. We will test the latter case in the next section.

Similar trends emerge for $T_3$ (green lines). In this case, we see a more pronounced decrease of $p_C(k)$ as $K$ increases. This dataset has few samples and the neighbourhood quickly starts including samples of other classes as $K$ grows.

Considering $T_2$ (blue lines), we observe similar results when $K\leq3$, with less polluted neighbourhoods when using \mae embeddings. Interestingly, as $K$ increases, \mae neighbourhoods become more polluted than those resulting from simple concatenation. That is, the concatenation of measurements could be sufficient to map the samples of the same class in the same areas of the metric space. Here, the compression performed by the \mae slightly confuses the neighbourhoods of samples, which, however, still exhibit high $p_C$. 

\vspace{3mm}
\textbf{Takeaway}: 
Although not trained for this purpose, the embeddings learned by the \mae exhibit nice properties pushing same-class samples closer to each other in the embedding space. This hints at their usefulness for downstream tasks but also opens the opportunity for successfully using small and simple models to solve these tasks, which we verify next. 
\section{Shallow Models}
\label{s:clustering}

One appealing property of good quality \mae encoded features is the possibility to use them for multiple downstream tasks, eventually with simple off-the-shelf ML algorithms. In this section, we show that the \mae encoded features allow simple shallow learning models to deliver good classification performance.   

\begin{figure}[t]
    \centering
    \includegraphics[width=.6\linewidth]{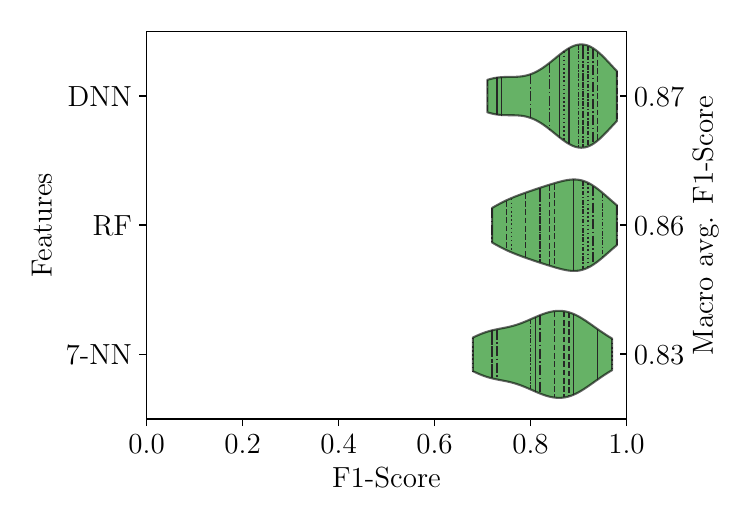}
    \caption{Classification performance in $T_1$ comparing deep and shallow models with the \mae encoded features.}
    \label{fig:task1_violin_sl}
\end{figure}

We repeat the supervised classification experiments of the previous sections using shallow learners with the \mae encoded features. We select two simplistic algorithms, namely (i) a Random Forest (RF) \citep{ho1995random} built using the Gini impurity index; and (ii) a K-Nearest-Neighbors (K-NN) classifier. For this experiment, we set $K=7$, similarly to \citep{gioacchini2021darkvec}. 

Results for the \texttt{MIRAGE} dataset are summarised in \cref{fig:task1_violin_sl}. Numbers in the right-hand axis of the figure report the macro average F1-Score. Notice in \cref{fig:task1_violin_sl} that the Deep Neural Network (DNN) model, RF, and K-NN results are all compatible. The more powerful DNN model still provides better figures, with 87\% macro average F1-Scores, against 86\% (83\%) for the RF (7-NN). 
Yet, the goodness of the shallow models with the \mae is worth noticing. 
That is, the compact representation obtained with the \mae allows good performance in this downstream task even with the more lightweight shallow models.  

\vspace{3mm}
\textbf{Takeaway}: 
Similar to other fields where representation learning is widely used~\citep{tian2020rethinking}, the condensed embedding extracted by \mae allows simpler models to be competitive with deep models, compared to the raw concatenation of features.

\section{Limitations}
\label{s:limitations}

Our work has limitations and unexplored questions which we summarise here. 
First, while we have searched for the best parameters for the \mae architecture we adopted, multiple other architectures could be employed for the same purpose. For example, we have listed in \cref{s:soa} different alternatives for the adaptation modules (\eg Graph Neural Network embeddings). These approaches could enhance the representations learned from network traffic. We have not tested such alternatives. Equally, finding the best NN architecture to serve as the Integration Module is still an open problem that we do not tackle in this work. 

Second, our work could be seen as a first step towards the creation of \textit{foundation network model}, i.e., models trained on vast amounts of network data to solve a wide range of problems and use cases. Given the robust embedding properties generated by \mae, we believe these representations hold significant promise in this direction, but other approaches are promising too. To arrive at such ambitious goals, we need to systematically extend the methodology to more use cases and datasets, which we do not face in this work. 

Third, practical traffic analysis problems are known to suffer from both temporal and spatial generalisation problems -- that is, models learned from a network in a specific time may not apply to other networks or different time frames (e.g., due to drift). Moreover, it is known that the traffic mix changes continuously, e.g., with the appearance of new applications and protocols, which change the observable entities and quantities. Here we do not investigate the impact of drift on the \mae.

Finally, we illustrate the application of the \mae using traffic classification problems. Traffic classification is still possible thanks to plain text fields present on protocol headers and payload as well as thanks to side channels, such as inter-packet times and packet sizes. The deployment of more techniques and protocols to improve online privacy would affect the \mae representations as for other traffic classification approaches.     

\section{Conclusions}
\label{s:conclusion}

In this paper, we proposed a generic architecture to learn vector representations from network data for traffic analysis. First, noticing a bi-modality in collected data where several entities co-exist with several measured quantities, we advocate the systematic use of NLP word embedding techniques to unsupervisedly extract rich entity features prone to machine learning models. We further propose to use one adaptation module per measurement type, trained with self-supervision to learn salient features from various quantities, before compressing all entity and quantity features by leveraging an auto-encoder. We objectively evaluate our approach and the quality of our learned embeddings on three different traffic classification downstream tasks, using open datasets for which we have labels and baseline results. 
Our results (i) confirm the usefulness of entity embeddings, often under-used in traffic analysis tasks, (ii) show that our compressed multi-modal embeddings do not lose information when fed to downstream tasks while streamlining the learning pipeline.
We open all our MAE models, together with our source code, in the hope of sparking the community for the search for better models and benchmarks, and for the eventual future development of foundation models for network applications.

\section*{Acknowledgements}
The research leading to these results has been partly funded by the Huawei R\&D Center (France), by the project SERICS (SEcurity and RIghts In the CyberSpace - PE00000014) under the MUR National Recovery and Resilience Plan funded by the European Union, as well as the ACRE (AI-Based Causality and Reasoning for Deceptive Assets - 2022EP2L7H) and xInternet (eXplainable Internet - 20225CETN9) projects - funded by European Union - Next Generation EU within the PRIN 2022 program (D.D. 104 - 02/02/2022 Ministero dell'Università e della Ricerca). This manuscript reflects only the authors' views and opinions and the Ministry cannot be considered responsible for them.

\bibliographystyle{plainnat}  
\bibliography{bibliography}

\appendix
\DeclareRobustCommand{\rchi}{{\mathpalette\irchi\relax}}
\newcommand{\irchi}[2]{\raisebox{\depth}{$#1\chi$}}

\section{Background on Adaptation Modules}
\label{appx:adaptation}

In this section, we provide some background about the architectures adopted as adaptation modules. Notice that we describe the architectures in each subsection independently, thus repeating the mathematical notation for clarity within its context.

\subsection{Convolutional Neural Network (CNN)}
\label{appx:cnn}

A Convolutional Neural Network~\cite{oshea2015introduction} operates through a series of layers designed to detect spatial hierarchies of features within input data. Mathematically, the operations conducted within a CNN involve convolutions, nonlinear activations, pooling, and fully connected layers, each contributing uniquely to the network's ability to understand intricate patterns and representations.

Consider a 2-D input sample (\eg an image) with $R$ rows and $C$ columns, $X \in \mathbb{R}^{R \times C}$. The core operation in a CNN is the convolution, which relies on a kernel $K \in \mathbb{R}^{n_1\times n_2}$.

The first stage of the convolution consists of obtaining a \emph{feature map} $F\in\mathbb{R}^{R-n_1 \times C-n_2}$ by applying the kernel $K$ on the input sample $X$. Hence, the $ij$-th entry of the feature map is obtained as:
$$
F[i,j]=(X\odot K)_{[i, j]}
$$
After the application of the kernel on the full input sample, the full convolution operation consists of applying an activation function $\phi_a$. Hence, the output of the convolution is obtained as $\rchi = \phi_a (F)$.

After convolution, CNNs usually employ pooling layers. They downsample the feature map, reducing dimention while retaining important information. Max pooling, for instance, selects the maximum value within a defined window. 

Pooling relies on two parameters: $p$, indicating the height and width of the pooling window, and the stride $s$, indicating the number of values the pooling window moves at each step. Hence, the $ij$-th component of the pooled sample $P^{p_1\times p_2}$ is defined as $P[i,j] = \max\left(\rchi[i:i+p\,,\,j:j+p]\right)$, where $p_1 = \frac{R-n_1-p}{s}+1$ and $p_2 = \frac{C-n_2-p}{s}+1$.

\begin{figure}[t]    \centering\includegraphics[width=.7\linewidth]{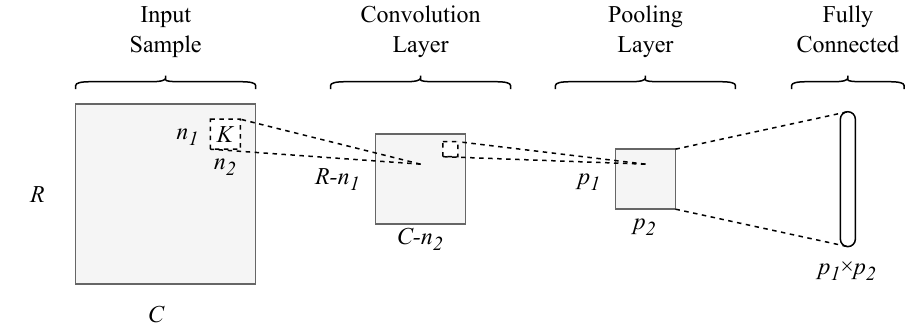}
    \caption{Convolutional Neural Network overview.}
    \label{fig:cnn}
\end{figure}

After multiple passes of convolution and pooling, the resulting output sample is then flattened and processed through one (or multiple) fully connected layers. In \cref{fig:cnn} we provide an overview of a single convolution-pooling step for one input sample.

\subsection{Word2Vec}
\label{appx:word2vec}

Word2Vec \cite{mikolov2013efficient, mikolov2013distributed} is a NLP technique based on artificial neural networks. It allows to map words (\textit{tokens}) of text sentences (\textit{corpora}) into a latent space as a real-valued array (the \textit{embedding}), such that words belonging to similar contexts have similar embedding.

The core element of the Word2Vec model is the \textit{context}. It is defined as the sequence of words surrounding the one for which the embedding must be generated.
The number of words to consider in the context is specified by the context window size $c$.
For example, by considering the sentence \textit{'Chicago is a great city'}, if the word \textit{'a'} is the target one and $c = 2$, the context for \textit{'a'} is the list of the $2$ previous and $2$ following words of \textit{'a'}:
$$
\text{\textit{'a'}}\rightarrow \text{(\textit{'Chicago', 'is', 'great', 'city'})}
$$

\begin{figure}[t]
    \centering\includegraphics[width=.7\linewidth]{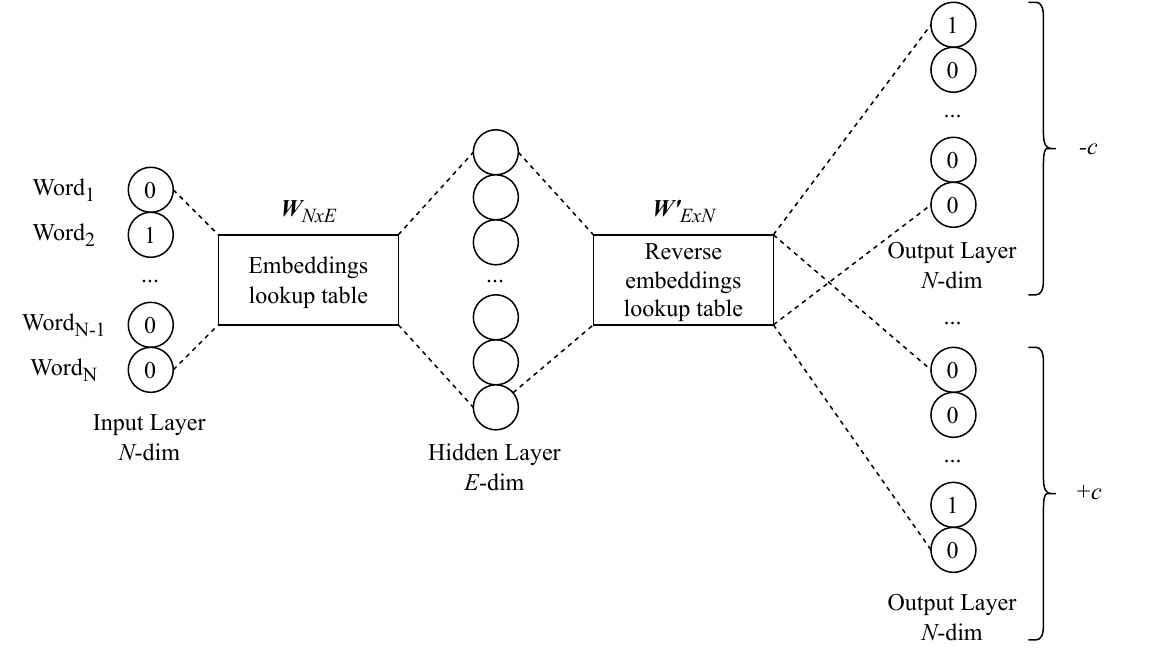}
    \caption{Word2Vec skipgram architecture.}
    \label{fig:skipgram}
\end{figure}

To generate the embedding, Word2Vec relies on two possible architectures: skip-grams and Continuous Bag Of Words (CBOW). Since we work with the skip-gram architecture, for the sake of simplicity we omit the description of CBOW. By considering a corpus with $N$ distinct words, the model aims to predict the probability of finding each one of the $N$ words within the context window of a given target word. In \cref{fig:skipgram} we report an overview of the skip-gram architecture. Each word of the sentences is fed as input to the model through a one-hot-encoded input layer. The $E$-dimensional hidden layer links all the $2c$ context words to the target one. After the model training, the embedding is obtained from the weights matrix $\mathbf{W} \in \mathbb{R}^{N\times E}$.
Each of the $i\in \{1, \dots, N\}$ entries of $\mathbf{W}$ is the embedding in $\mathbb{R}^{1\times E}$ associated to the \textit{i}-th word.

\subsection{Gated Recurrent Unit (GRU)}
\label{appx:gru}

Gated Recurrent Units (GRUs)~\cite{cho2014learning} is a particular type of Recurrent Neural Network (RNN), a family of neural networks designed for handling sequences of data. Unlike traditional models, GRUs have built-in mechanisms that help them decide what information to remember or forget, making them great for understanding patterns in sequences like text, audio, or time series data.

The GRU maintains a \textit{hidden state} to keep track of past information and merges it with new incoming data producing a latent representation of the input data.
Formally, let's consider a sequence of data extracted from a dynamic system $\{X^t\}_{t=0}^T$, of length $T$.
At each time interval $t$, the GRU receives as input a sample $x^{t} \in \mathbb{R}^{F}$, where $F$ is the number of features,\footnote{Notice that the GRU can process a matrix sized $N\times F$, where $N$ is the number of samples. For the sake of simplicity, we here report the formulation for only one sample. The generalisation is straightforward}, and the previous hidden state $h_u^{t-1} \in \mathbb{R}^{E}$, where $E$ is the size of the output vector provided by the GRU. It then transforms the two inputs into the new hidden state $h^{t} \in \mathbb{R}^{E}$ through two gates, the \textit{update} and \textit{reset} gate. 
The \textit{update} gate (output $c_t\in \mathbb{R}^{E}$) is used to balance the influence of the past hidden state and the newly acquired information on the output.
In this way, the GRU provides a direct connection between initial input and output which helps modelling long-term dependencies and improves gradient flow. The \textit{reset} gate $r_t\in \mathbb{R}^{E}$ modulates the contribution of the past hidden state to the newly added information. 

\begin{figure}[t]
    \centering\includegraphics[width=.4\linewidth]{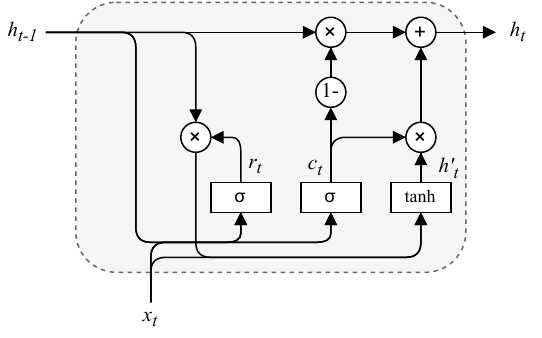}
    \caption{Gated Recurrent Unit architecture.}
    \label{fig:gru}
\end{figure}

Formally, 
$$
    c_t=\sigma(W^cx_t+U^ch_{t-1})
$$
$$
    r_t=\sigma(W^rx_t+U^rh_{t-1})
$$
$$
    h'_t=\tanh\left(Wx_t+U(r_t\odot h_{t-1})\right)
$$
$$
    h_t=c_t\odot h_{t-1}+(1-c_t)\odot h'_t
$$
where $W^c, W^r, W \in \mathbb{R}^{E\times F}$ and $U^c, U^r, U \in \mathbb{R}^{E\times E}$ are learnable matrices, $\odot$ is the element-wise product and $\sigma$ is the sigmoid function. For the sake of completeness, we report in \cref{fig:gru} the architecture of a single GRU cell.
\begin{table}[!t]
\centering
\footnotesize
\caption{Architectures of the adaptors and encoders}
\begin{tabular}{lllcl}
\toprule
\textbf{Feature} & \textbf{Stage} & \textbf{Layer Type} & \textbf{Units} & \textbf{Note} \\
\midrule
\multirow{9}{*}{Subnet} & \multirow{4}{*}{Encoding} & GRU & 32 & Return sequences = True \\
 &  & GRU & 32 & Return sequences = False \\
 &  & Dense & 64 &  \\
 &  & Dense & $l_1$ & By default $l_1$=32 \\ \cmidrule(lr){2-5}
 & \multirow{5}{*}{Decoding} & Dense & $l_1$ & By default $l_1$=32 \\
 &  & Dense & 64 &  \\
 &  & RepeatVector & -- & 4 times \\
 &  & GRU & 32 & Return sequences = True \\
 &  & Dense & 1 & TimeDistributed \\
 \midrule
\multirow{11}{*}{Payload} & \multirow{6}{*}{Encoding} & Embedding & 64 & Vocabulary size: 257 \\
 &  & Masking & -- & Padding value:0 \\
 &  & GRU & 64 & Return sequences = True \\
 &  & GRU & 32 & Return sequences = False \\
 &  & Dense & 64 &  \\
 &  & Dense & $l_1$ & By default $l_1$=32 \\
 \cmidrule(lr){2-5}
 & \multirow{5}{*}{Decoding} & Dense & $l_1$ & By default $l_1$=32 \\
 &  & Dense & 64 &  \\
 &  & RepeatVector & -- & 32 times \\
 &  & GRU & 64 & Return sequences = True \\
 &  & Dense & 1 & TimeDistributed \\
 \midrule
\multirow{11}{*}{Sequences} & \multirow{4}{*}{Encoding} & Conv1D & -- & Filters: 32; Kernel size: 3 \\
 &  & MaxPooling1D & -- & Pool size: 2 \\
 &  & Flatten & -- &  \\
 &  & Dense & $l_1$ & By default $l_1$=32 \\
 \cmidrule(lr){2-5}
 & \multirow{7}{*}{Decoding} & Dense & $l_1$ & By default $l_1$=32 \\
 &  & Dense & 480 &  \\
 &  & Conv1D & -- & Filters: 32; Kernel size: 3 \\
 &  & UpSampling1D & -- & Size=2 \\
 &  & Conv1D & -- & Filters: 4; Kernel size: 3 \\
 &  & UpSampling1D & -- & Size=2 \\
 &  & Dense & 32x4 &  \\
  \midrule
\begin{tabular}[c]{@{}l@{}}IP address, Ports\\ Statistics\end{tabular} & \begin{tabular}[c]{@{}l@{}}Encoding,\\ Decoding\end{tabular} & Dense & $l_1$ & By default $l_1$=32\\
 \bottomrule
\end{tabular}
\label{tab:architecture}
\end{table}

\section{Architectures of the Adaptation Modules}
\label{appx:architectures}

We now complement the details of the architectures and implementation of the adaptation modules.
In \cref{tab:architecture} we report details about the implementation of the adaptation modules. For the sake of simplicity, we report the layer names provided by the Keras API~\citep{teamnodatekeras}. We use ReLU as the activation function for all hidden layers.
\section{\mae parameter tuning}
\label{s:complexity}

\begin{figure}[!t]
    \centering
    \begin{subfigure}[b]{.48\textwidth}
        \centering
        \includegraphics[width=\linewidth]{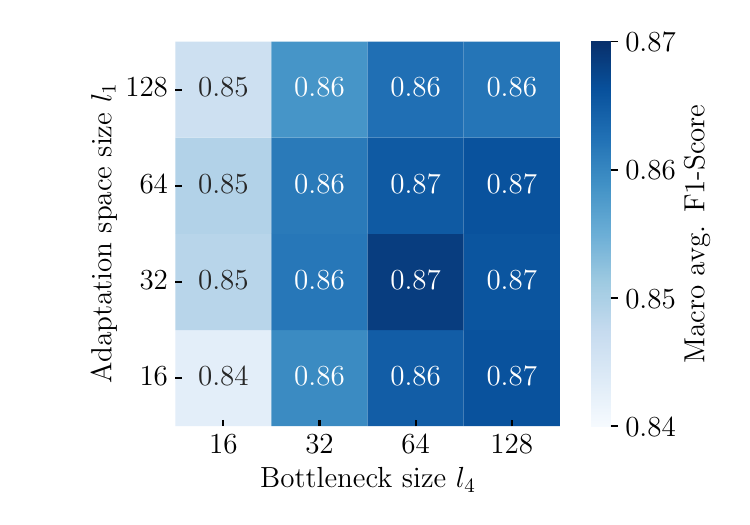}
        \caption{Macro F1-Score.}
        \label{fig:gs}
    \end{subfigure}
    \begin{subfigure}[b]{.48\textwidth}
        \centering
        \includegraphics[width=\linewidth]{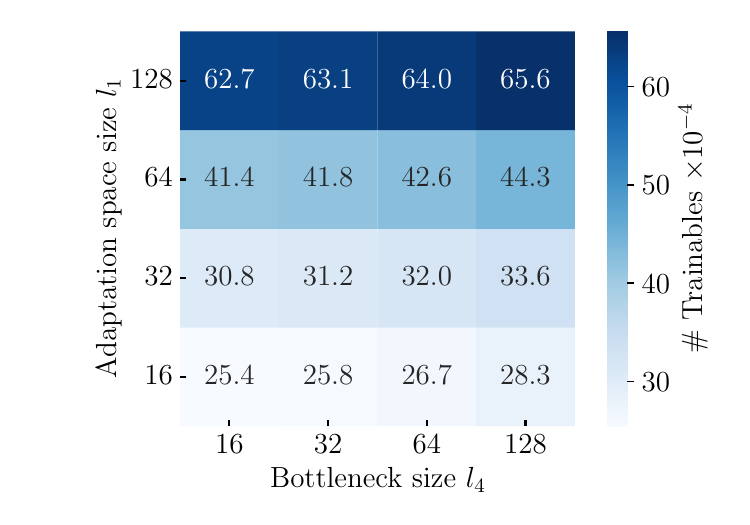}
        \caption{Size of the neural network (encoder).}
        \label{fig:params}
    \end{subfigure}
    \caption{Impact of parameter choices.}
    \label{fig:gridsearch}
\end{figure}

We have seen in \cref{s:rl} that the \mae is defined by multiple parameters (\eg $l_1, \ldots, l_4$), which are summarised in \cref{tab:architecture}. When designing the \mae we investigate the impact of these parameters using classic grid-searching methodologies. Here we provide a summary of the results. 

\cref{fig:gridsearch} summarises the impact of the adaptation space size $l_1$ and bottleneck size $l_4$ for $T_2$. \cref{fig:gs} shows how the macro average F1-Scores are impacted by these parameters, while \cref{fig:params} provides the size of the obtained autoencoder, defined as the number of internal weights and biases (\eg trainables). We see that varying the size of $l1$ and $l4$ has a minor impact on the average F1-Scores. The best result ($l1=32$ and $l4=64$) is 87\% for $T_2$, which is only marginally better than the worst-tested cases. \cref{fig:params} instead shows that these parameters have a large impact on the size of the autoencoder, thus calling for moderation when picking these dimensions.\footnote{Training and validating one model with the \texttt{DARKNET} dataset (44k samples) takes around 5 minutes, using 150 epochs in a single Nvidia Tesla V100 GPU with 16GB of memory.} 

Overall, we have not identified a major impact on F1-Scores when varying these parameters in all three downstream tasks. As our goal is to produce a representation that is generic and reusable for multiple problems, we argue that the \mae \emph{should not} be over-optimised for a single downstream task. Indeed, generic parameters that fit multiple tasks well, as above, should be preferred.

\end{document}